\documentclass[10pt,twocolumn,letterpaper]{article}

\usepackage{3dv}
\usepackage{times}
\usepackage{epsfig}
\usepackage{graphicx}
\usepackage{amsmath}
\usepackage{amssymb}
\usepackage{enumitem}
\usepackage[dvipsnames]{xcolor}

\newcommand{\old}[1]{}

\usepackage{booktabs}
\usepackage{xr-hyper}

\usepackage[pagebackref=true,breaklinks=true,colorlinks,bookmarks=false]{hyperref}

\threedvfinalcopy 

\def\threedvPaperID{258} 

\ifthreedvfinal\fi
\begin{document}

\title{Deep Sketch-Based Modeling: Tips and Tricks}

\author{Yue Zhong$^{1,3}$
\and
Yulia Gryaditskaya$^{1,2}$
\and
Honggang Zhang$^{3}$
\and
Yi-Zhe Song$^{1,2}$
\and
\\
$^{1}$SketchX, CVSSP, University of Surrey $^{2}$iFlyTek-Surrey Joint Research Centre on Artificial Intelligence \\
$^{3}$ Beijing University of Posts and Telecommunications\\
}

\maketitle

\begin{abstract}
Deep image-based modeling received lots of attention in recent years, yet the parallel problem of sketch-based modeling has only been briefly studied, often as a potential application. In this work, for the first time, we identify the main differences between sketch and image inputs: (i) style variance, (ii) imprecise perspective, and (iii) sparsity.
We discuss why each of these differences can pose a challenge, and even make a certain class of image-based methods inapplicable. We study alternative solutions to address each of the difference. By doing so, we drive out a few important insights: (i) sparsity commonly results in an incorrect prediction of foreground versus background, (ii) diversity of human styles, if not taken into account, can lead to very poor generalization properties, and finally (iii) unless a dedicated sketching interface is used, one can not expect sketches to match a perspective of a fixed viewpoint. Finally, we compare a set of representative deep single-image modeling solutions and show how their performance can be improved to tackle sketch input by taking into consideration the identified critical differences. 

\end{abstract}

\section{Introduction}

The challenge of being able to obtain a 3D model from a single sketch has been intriguing research for decades. Typically, proposed methods make assumptions on the type of the input \cite{xu2014true2form} or restrict the users to a specific user interface \cite{bae2008ilovesketch, igarashi2006teddy, schmidt2009analytic}. Being an under-constrained problem, for which it is hard to devise a reliable set of heuristics, it naturally asks for deep learning-based methods. 

Nevertheless, despite the recent surge of image-based reconstruction, deep sketch-based modeling remains largely overlooked. In this work, we evaluate the applicability of state-of-the-art deep single RGB image methods to the sketch-based modeling problem. We discuss the main challenges and differences between sketch and image inputs -- style variance between humans, imprecise perspective, and sparsity -- and propose universal solutions to increase the robustness of existing methods on a sketch input. 

The first challenge comes from style differences, i.e., each person carries a unique sketching style.
To address this problem, we generate three synthetic datasets: naive, stylized, and one where the style is unified by an additional image processing network. The naive dataset represents rendering style with a uniform line width, commonly used in sketch-based reconstruction papers, where the lines are obtained from 2D images or via non-photorealistic rendering from 3D models. In this work, we rely on the latter.
The stylized dataset is designed to capture the diversity of human sketching styles. The strategy of the style-unifying image translation network was proposed in \cite{wang20203d}, and was shown to be efficient on doodle sketches. We aim at more detailed sketches and show that if the sketching style is within an expected variance on line widths and over-sketching then training on the proposed stylized dataset results in more accurate reconstructions compared to a style-unifying network.

It is common for deep single image methods to train and test their models on a predefined set of viewpoints \cite{lun20173d}. Nevertheless, it was observed by Gryaditskaya et al.~\cite{gryaditskaya2019opensketch} that even professional designers, when asked to sketch from a given viewpoint, produce sketches with large angular deviations from the set viewpoint. We thus create a dataset by generating for each shape 48 viewpoints, where 8 viewpoints are fixed and 5 additional viewpoints for each viewpoint are randomly sampled from a normal distribution with the mean matching the parameters one of the fixed viewpoints. To explicitly account for the variation of styles and viewpoints, we aim at learning style- and viewpoint-invariant shape representation by proposing a regression loss that encourages the correlation of Chamfer distances and dot product distances in the feature space.

The final challenge in deep sketch-based reconstruction comes from the difficulties in distinguishing the foreground from the background, due to the sparsity of sketch lines. To alleviate this problem we suggest a simple framework, where the user can provide a few sparse labels, and the network is able to propagate these labels to robustly predict shape foreground binary mask, which is then passed to a 3D shape reconstruction network alongside the input sketch. 

In summary, we propose the following contributions:
\begin{itemize}
    \item We identify key differences between images and sketches, and discuss the challenges it imposes on deep-reconstruction methods.
    \item We compare alternative strategies to handle human sketching styles variations.
    \item We adopt the regression loss to learn style- and viewpoint-invariant sketch embedding.
    \item Finally, we propose to use an auxiliary network that learns to predict foreground mask from the input sketch, and supports user sparse labels when necessary. We demonstrate how such mask can be incorporated as an input to a reconstruction network and allows to account for sparsity of input sketches.
\end{itemize}

\section{Related work}
For a general overview of existing sketch-based reconstruction methods, please refer to a recent survey by Bonnici et al.~\cite{bonnici2019sketch}. In this section, we give an overview of deep single image reconstruction methods and discuss in detail existing deep methods for sketch-based reconstruction.

\paragraph{Single RGB image to 3D.}

\emph{Multi-view.} 
Tatarchenko et al.~\cite{tatarchenko2016multi} exploit an encoder-decoder convolutional network and predict unseen viewpoints and depth maps, which are consequently fused into a dense point cloud.  Notably, they train with arbitrary viewpoints and lighting conditions generating the training data on-the-fly.
The architecture of Yao et al.~\cite{yao2019front2back} is based on a conditional adversarial network. They first predict silhouettes, normal and depth maps for an input view, they then analyse the shape for the symmetry and if one is detected, they use the reflected view to complement the initial view to predict the back view of the shape. The disadvantage of such methods is a heavy processing required to obtain the final mesh.

\emph{Voxels.} Voxel shape representation \cite{wu2016learning, choy20163d, girdhar2016learning,liao2018deep, wu2018learning} has regular structure and thus allows to adopt state-of-the-art techniques for 2D images. The disadvantages of such methods are large memory footprint and low-resolution results.

\emph{Implicit representation.} Recently a number of works were proposed which use implicit functions to represent a shape \cite{mescheder2019occupancy,chen2019learning,chen2020bsp}. Some works learn \cite{mescheder2019occupancy,chen2019learning} to predict for each point if it is free or occupied. The disadvantage of these methods is that in order to reconstruct the 3D shape the heavy processing is still required. Chen et al.~\cite{chen2020bsp} directly outputs an approximation of a surface, and handles well sharp shape features, what though comes at the cost of poor reconstruction of curved surfaces. 

\emph{Deformable mesh patches.}
Groueix et al.\cite{groueix2018papier} proposed to approximate the surface locally by mapping a set of squares to a 3D shape. This approach produces not closed meshes that can have holes and self-intersections.

\emph{Image to mesh.}
A pioneering work on a direct image to mesh translation \cite{wang2018pixel2mesh} introduced a graph-based convolutional neural network limited to 3D meshes with genus 0, that predicts the deformation of an ellipsoid. It, however, relies on the known camera intrinsic parameters.
Pan et al.~\cite{pan2019deep} lift this limitation by introducing an additional module for errors predictions and mesh faces removal step.

\emph{Point clouds.}
There is a number of works which target point clouds shape representation, since it is a native representation for the scanned shapes and scenes. Fan et al.~\cite{fan2017point} proposed a conditional generative encoder-decoder network for a single image point cloud reconstruction. Achlioptas et al.~\cite{achlioptas2018learning} studied the evaluation metrics for point-clouds comparison. Gadelha et al.~\cite{gadelha2018multiresolution} proposed a multi-resolution decoder that improves information flow on multiple scales, and improves the quality of the generated shapes. Yang et al.~\cite{yang2018foldingnet} proposed an auto-encoder for unsupervised learning on point clouds. 

\emph{Unsupervised methods.} 
Neural rendering enables unsupervised training for single image 3D reconstruction \cite{loper2014opendr, kato2018neural, liu2019soft}, which is achieved by comparing the estimated shape silhouette with an input image.

\paragraph{Deep sketch-based modeling}

\emph{Voxel-based.} Delanoy et al.~\cite{delanoy20183d} proposed a U-Net \cite{ronneberger2015u} based architecture, where they encode 3D shape in a voxel representation and estimate the probability of each voxel to be occupied. Their method exploits the dedicated sketching interface, and their multi-view sketching shape update strategy relies on the known perspective camera-parameters.
Jin et al.~\cite{jin2020contour} learn the embedding of the shape given silhouettes of the 3D shape from the front, side and top views. While it only exploits the information contained in the shape silhouettes, it proposes an interesting idea for single sketch modeling of retrieving the two additional views in the embedded space, prior to 3D reconstruction. They convert the voxelized shape representation to a mesh by a marching cube algorithm \cite{lorensen1987marching}.

\emph{Patch-based.} Smirnov et al.~\cite{smirnov2019deep} proposed a new shape representation as an assembly of Coons patches, where the main goal is to obtain the representation, in an end-to-end manner, that can be easily manipulated by designers. We do not evaluate their method due to unavailability 
of the code. 

\emph{Point cloud.} Wang et al.~\cite{wang20203d} adopted \cite{fan2017point} for sketch-based 3D reconstruction by proposing an additional image translation network that aims at sketching style standardization to account for the variability of sketching styles. In addition, to support arbitrary viewpoints they utilize the shape transformation module from \cite{qi2017pointnet}. Similarly, to this work we test here the importance of style-standardization module and explore alternative strategies to enable style- and viewpoint-invariance.  

\emph{Single to Multi-view methods.} Lun et al.~\cite{lun20173d} predict normal and depth maps as seen from 12 viewpoints, that are fused to a dense point cloud. Li et al.~\cite{li2018robust} target free form surfaces and introduced the intermediate layer that predicts dense curvature directions. The method supports sparse labels for depth maps and curvature hints for strokes. It makes assumptions on line rendering: e.g., the silhouette lines are assumed to be sketched in black and other lines in grey. Compared to them we do not make any assumptions on sketching style, and our method offers support for sparse labeling of foreground versus background, a much simple task for a human.

\section{Datasets}
In this section, we describe the selected shapes, viewpoints, and the rendering settings for our training and test data.

\subsection{Synthetic datasets}
\begin{figure}[htb]
\begin{center}
   \includegraphics[width=\linewidth]{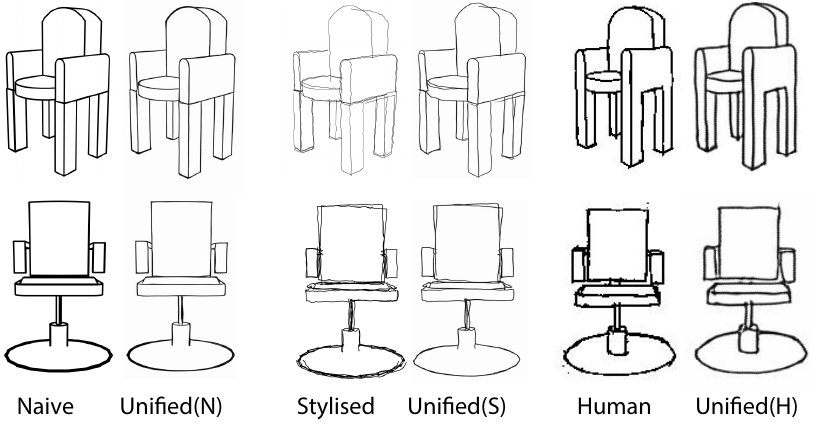}
\end{center}
   \caption{Example sketches from our naive, stylised and style-unified synthetic datasets, as well as sketches from ProSketch3D dataset of human sketches. }
\label{fig:synthetic_dataset}
\vspace{-10pt}
\end{figure}

We generate three datasets with distinctive styles, which we refer to as naive, stylized and style-unified (Figure \ref{fig:synthetic_dataset}).

\paragraph{Naive sketch.} Naive sketch denotes synthetic sketch generated from a reference 3D model, using silhouettes and creases rendering, with a uniform stroke width, which we set to $2.5$. We render such $256\times256$ sketches using Blender Freestyle. Such types of sketches are clean and perspectively accurate, and thus the reconstruction results on them can achieve higher accuracy. Nevertheless, such sketches differ from human sketches that commonly exhibit perspective and mechanical inaccuracies, as well as over-sketching. Despite the existence of multiple recent solutions that aim at converting such rough sketches to cleaner ones \cite{simo2015discriminative,bessmeltsev2019vectorization}, these solutions are prone to interpretation mistakes. We thus propose two additional datasets: the one that aims at directly mimicking human sketching styles, and the one that represents sketches with the style unified by an addition sketch processing network. 


\paragraph{Stylized sketch.} 
To obtain a stylized dataset we apply a set of random global and local deformations to each stroke of a naive sketch, exploiting the `\emph{svg\_disturber}' script from the open-source library\footnote{https://gitlab.inria.fr/D3/contour-detect/\-/blob/master/svg\_tools/svg\_disturber.py}. To obtain a sketch in a vector format we use the Blender SVG exporter. 

The global stroke deformation consists of stroke rotation, scaling and translation.
The rotation angle is randomly sampled from $[0, 2.5^\circ]$.
We allow stroke global scaling that does not preserve stroke aspect ratio, where the scale factor is randomly sampled from $[0.9, 1.1]$. Finally, the translation vector is randomly sampled from the disk with a $2.5$ radius.
We enable coherent local noise, where the offset is sampled randomly from $[0, 1.3]$ interval. In addition, we enable over-sketching, meaning that the stroke under global a local deformation is traced several times. We allow the width of the strokes to be traced at most two times. We allow the strokes width within one sketch to vary with its width value being randomly sampled from the normal distribution with mean set to 2.5 and variance equal to 1.5.

\paragraph{Style-unified sketch.}
Inspired by recent work \cite{wang20203d}, which deploys an image translation network prior to a 3D reconstruction, and variety of methods aiming at sketch consolidation/simplification/beautification \cite{simo2015discriminative,bessmeltsev2019vectorization,liu2018strokeaggregator}, we generate an additional dataset by passing stylized sketches through the fully convolutional sketch simplifying neural network \cite{simo2015discriminative} (Figure \ref{fig:synthetic_dataset}). The output sketches have unified stroke appearance, what facilitates the inference when each unseen sketch is first passed through the same simplifying network.
\subsection{Selected shapes}
Most of our experiments are conducted on the models from a chair category of the ShapeNetCore dataset\footnote{https://www.shapenet.org/}, complemented by two additional categories: planes and lamps. We selected these categories guided by the next principles:

\noindent \emph{Easy to sketch.} 3D shape should have a simple structure and should be easy to draw for a human.\\
\emph{Generality.}  We focus on common categories, that are well familiar to humans. For instance, chairs are common for everyday life, while rifles is an example of a more specific category.\\
\emph{View differentiability.} Each shape is expected to have a distinct appearance at distinct viewpoints.\\
\emph{Shape genius higher than 1.} We select chairs, which contain 6778 models, as our main test category, due to variability in level of details and topologies.\\
\emph{Large inter-category variance.} We complement chairs with the two other common categories, where airplanes represent shapes with large variability of surface curvatures (4045models) and lamps contain fine-scaled details (2318 models).

\subsection{Selected viewpoints}
\label{sec:viewpoints}
For each 3D shape, we first generate sketches for 8 base viewpoints, where the camera elevation is set to 10 degree to imitate human perspective in the real world, and azimuth takes values equidistantly sampled from 0 to 360 degree.
For each base viewpoint, we generate 5 additional viewpoints by sampling the camera elevation and azimuth angles from normal distributions with mean matching the elevation and azimuth of one of the base viewpoints. 
We set variance to $7$ degrees.
To avoid viewpoints too similar to base viewpoints or viewpoints deviating too much from them, we add the lower and upper thresholds of $5$ and $15$ degrees, respectively.
The distance between a virtual camera and a 3D shape is randomly sampled from the normal distribution with mean set to 1.5 in the range [1.4,1.6]. We use a perspective camera for all viewpoints.

\subsection{Human sketch dataset}
\label{sec:human_dataset}
We exploit a ProSketch3D\footnote{http://sketchx.ai/downloads/} dataset of professional human sketches of chairs from ShapeNetCore dataset as an additional test set. ProSketch3D dataset was collected using a ISKN Slate 2\footnote{https://www.iskn.co/uk/} digital drawing tablet. This dataset contains both sketches in PNG and SVG formats. This dataset was deliberately designed to contain little style variations and perspective inaccuracies by carefully selecting the participants and providing them with a drawing example, and letting to sketch over the reference viewpoint. Each shape has three viewpoints: front, side and $45^\circ$. Yet, it contains small mechanical inaccuracies non-present in naive sketches and thus presents an interesting test case. We use their provided PNG sketches, where the line width rendering roughly matches the 2.5 line width of the the sketches in our naive dataset. 

\section{Reconstruction Baselines}
We train and test a number of single image reconstruction baselines with view-based and volume-based shape representation on our synthetic and human datasets.

\label{sec:baselines}
\subsection{View-based}
\label{sec:baseline_ShapeMVD}
 We selected a recent approach by Lun et al.~\cite{lun20173d}, to which we further refer as \emph{ShapeMVD}, dedicated to depth and normal maps predictions from input sketches, which are then merged into a dense point cloud. The mesh is obtained by Poisson Surface Reconstruction \cite{nealen2005sketch}. The method allows fine-tuning if the camera parameters are known, by rendering the reconstructed mesh and smoothly deforming the mesh so that the rendered contour matches the input sketches. We do not use this step, since in a general setting the viewpoint for a human sketch is unknown. In the original paper, the model was trained with two input sketches: the two orthographic projections of the shape from the front and from the side with clean uniform line-rendering. Here we retrain their architecture with our datasets, using only a single perspective sketch as an input.

\subsection{Volume-based}
We consider here a number of different state-of-the-art methods and select one to two representative methods for different shape representations.

\paragraph{Mesh}
Since being able to directly obtain a mesh from the input sketch is highly desirable, we evaluate two baselines that directly predict a mesh: \emph{Pixel2Mesh} \cite{wang2018pixel2mesh}, which is commonly used as a comparison baseline and an approach by Pan et al.~\cite{pan2019deep} (which we refer to as \emph{TMNet}) due to its ability to handle shapes of arbitrary genus.
The TMNet method consists of three main units: shape deformation estimation, errors predictions and boundary topology refinement. 
The error-prediction module is trained to predict the error distances between the predicted and ground-truth shapes, given an input image feature vector. The errors prediction is used for face pruning step to modify shape connectivity. This step allows capturing shape details much better than previous methods.
The boundary refinement unit enforces smoother boundaries, emerged after the faces pruning step.
It estimates the displacement of vertices rather than the absolute positions of the vertices, which enables a more efficient training.

\paragraph{Point cloud}
As a network that works with point-based shape representation, we exploit \emph{PSGN} \cite{fan2017point}. We use their vanilla encoder-decoder architecture with the Chamfer distance as a reconstruction loss.

\paragraph{Space occupancy}
Mescheder et al.~\cite{mescheder2019occupancy} proposed an alternative approach of learning continuous prediction of point occupancy, referred to as \emph{OccNet}. The 3D mesh reconstruction is obtained through iterative grid subdivision, where the grid cells which have neighboring grid cells with different occupancy labels are refined iteratively.
The shape iso-surface is extracted with Marching Cubes algorithm \cite{lorensen1987marching}, followed by the refinement step that
optimizes for normals smoothness.

\paragraph{Voxels}
As a network representative of the networks operating with a regular volumetric grid, we use \emph{3D-R2N2} \cite{choy20163d}. 

\section{Foreground binary mask}
\label{sec:mask}
\begin{figure*}[t]
\vspace{-8pt}
\begin{center}
  \includegraphics[width=\textwidth]{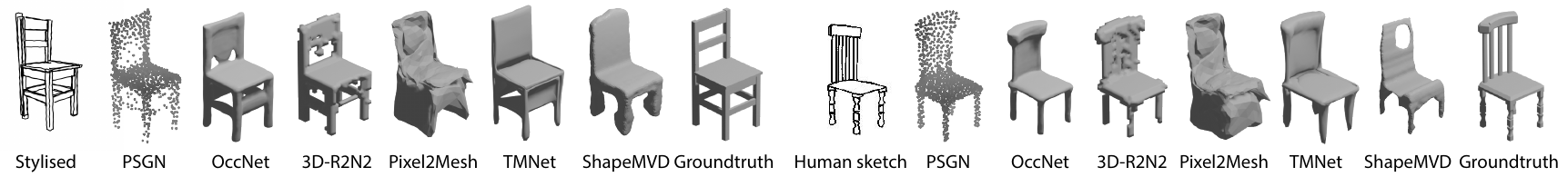}
\end{center}
  \caption{Visualization of reconstruction results with different baselines on test synthetic stylized dataset and human sketches.}
\label{fig:mask-sketch}
\vspace{-5pt}
\end{figure*}

Due to the sparsity of information in sketch image, single image reconstruction methods often can not reliably distinguish foreground from the background, what is demonstrated in Figure \ref{fig:mask}. To alleviate this problem we exploit image translation networks and the idea of interactive sparse user labeling \cite{zhang2017real, su2018interactive} to first predict foreground binary mask. Following these works, we leverage UNet image translation architecture. We train the network with sparse ground-truth labels marking both foreground and background with the following loss function:
\begin{multline}
\mathcal{L}_{c G A N}(G, D)=\mathbb{E}_{x, y}[\log D(x, y)]+\\
 \mathbb{E}_{x, \tilde{y}}[\log (1-D(x, G(x, \tilde{y}))] - \lambda_1 \mathcal{L}_{L1},
\end{multline}
where $x$ is a sketch image from the sketch domain $p_s$, y is a foreground binary mask from the domain $p_m$ and $\tilde{y}$ is a predicted foreground binary mask. $D(\cdot)$ is a discriminator conditioned on an input sketch. $\mathcal{L}_{L1}$ is an $L1$ loss that minimizes per pixel distances between the ground-truth foreground mask $y$ and predicted $\tilde{y}$. 
At the training stage, we sample sparse labels independently from both foreground and background according to a geometric distribution with a success probability set to $1/8$.  We follow the sampling strategy proposed in \cite{zhang2017real} and select the position of a label by sampling a 2D Gaussian distribution with the mean $\mu = 0.5[H,W]$ and covariance matrix $\Sigma = diag([(0.5H)^2, (0.5W)^2])$.

For OccNet, PSGN and 3D-R2N2 baseline methods we build additional architectures that takes both a sketch and a predicted binary foreground mask as an input. The mask and the sketch are separately passed through two convolutional layers. We then concatenate the outputs of each of the two branches and pass them to an encoder (see supplemental for the details).

\section{Embedded space}
\label{sec:regression_loss}
We propose a training strategy with an additional regression loss that aims at obtaining sketch embedding invariant to viewpoint and style. To achieve this goal we adopt a regression loss introduced in \cite{uy2020deformation} for deformation-aware 3D model embedding and retrieval. 

Our goal is to establish an order in the embedded space such that if the distance between 3D shape $\mathbf{A}$ and $\mathbf{B}$ is larger than the distance between shape $\mathbf{A}$ and $\mathbf{C}$, then the distance between the embeddings of sketches of shapes $\mathbf{A}$ and $\mathbf{B}$ is also larger than the distance between the embeddings of sketches of shapes $\mathbf{A}$ and $\mathbf{C}$.

Let $\mathbf{A} \subset \mathbf{X}$ and $\mathbf{B} \subset \mathbf{X}$ be two shapes and $f_{\mathbf{S_A}}$ and $f_{\mathbf{S_B}}$ be the embedding of the sketches $\mathbf{S_A}$ and $\mathbf{S_B}$ of the corresponding shapes, where $\mathbf{X}$ is a set of all 3D shapes.
As the distance between two shapes we exploit Chamfer distance (see Section \ref{sec:measures} for a precise definition), denoted as $d_{CD}(\mathbf{A},\mathbf{B})$, and as a distance in the embedded space $\delta(f_\mathbf{S_A},f_\mathbf{S_B})$ we use a dot product. We normalize the feature space to a unit hyper-sphere prior to a distance computation. During the training stage, for each iteration, we first convert all the distances to a form of probability distributions.
We convert Chamfer distances as follows:
\begin{equation}
p(\mathbf{A}, \mathbf{B}) = \frac{\exp \left(-d_{CD}^{2}(\mathbf{A}, \mathbf{B}) / 2 \sigma_{\mathbf{A}}^{2}\right)}
{\sum_{\mathbf{B}^{\prime} \in \mathbf{X}^{\prime}} \exp \left(-d_{CD}^{2}(\mathbf{A}, \mathbf{B}^{\prime}) / 2 \sigma_{\mathbf{A}}^{2}\right)},
\end{equation}
where $\mathbf{X}^{\prime} \subset \mathbf{X}$ is a randomly sampled fixed size subset for each iteration. $\sigma_{A}$ is a pre-computed constant for each 3D shape $\mathbf{A}$, which is calculated according to a three-sigma rule $\sigma_{A} = \frac{0.997}{3}d_{{C D}_{max}}$, where $d_{{C D}_{max}}$ is the maximum Chamfer distance to any of 3D shapes in the dataset, approximated by evaluating distances on a sufficiently large subset of $\mathbf{X}$. This non-linear mapping allows to more accurately learn an embedding for shapes which are similar to each other. Similarly, the distance in the embedded space $\delta(f_\mathbf{S_A},f_\mathbf{S_B})$ is converted into a probability distribution: 
\begin{equation}
\hat{p}(f_\mathbf{S_A}, f_\mathbf{S_B})=\frac{\exp(f_\mathbf{S_A} \cdot f_\mathbf{S_B})}
{\sum_{\mathbf{B}^{\prime} \in \mathbf{X}^{\prime}} \exp(f_\mathbf{S_A} \cdot f_\mathbf{S_{B^\prime}})}.
\end{equation}

Finally, the regression loss is defined as the $l 1$-distance of the two probability distributions:
\begin{equation}
\mathcal{L}_{R}\left( \mathbf{X}^{\prime}\right)=\frac{1}{\mathbf{X}^{\prime}} \sum_{\mathbf{A} \in \mathbf{X}^{\prime}}|\hat{p}(f_\mathbf{S_A}, f_\mathbf{S_B})-p(\mathbf{A}, \mathbf{B})|.
\end{equation}

While we rely on Chamfer distance, some alternative measures can be used, such as the ones summarized in Section \ref{sec:measures}, and used for the evaluation.

\section{Experiments}
We evaluate each of the proposed components with respect to identified challenges when dealing with sketch input: (i) style variance (Section \ref{sec:variance}), (ii) sparsity (Section \ref{sec:sparsity}) and (iii) view-inaccuracies (Section \ref{sec:view}).

\subsection{Test datasets viewpoints}
\label{sec:test_viewpoints}
We randomly select 500 shapes from each category as test datasets. For the chair category, we select the same split as was used to collect the ProSketch3D dataset (Section \ref{sec:human_dataset}). For a fair, yet tractable evaluation (inference for some baselines is quite costly), for each 3D shape in the synthetic test dataset, we randomly chose one of 48 viewpoints. For the human dataset, for each shape, we, similarly, randomly select one of the three available viewpoints.
The random sampling is done once, thus the test datasets are exactly the same for all the evaluated models. 

\subsection{Evaluation metrics}
\label{sec:measures}
We use four evaluation metrics to compare the predicted 3D shapes to the reference shapes: Chamfer distance \cite{fan2017point}, Earth mover's distance \cite{fan2017point}
and F-Score \cite{tatarchenko2019single}. 

\paragraph{Chamfer distance (CD).}
Chamfer distance measures the squared distance between each point in one point set to its nearest neighbor in the other set:

\begin{equation*}
 d_{CD}\left(S_{1}, S_{2}\right)=\sum_{x \in S_{1}} \min _{y \in S_{2}}\|x-y\|_{2}^{2}+\sum_{y \in S_{2}} \min _{x \in S_{1}}\|x-y\|_{2}^{2},
\end{equation*}
where $S_{1} \subset \mathbb{R}, S_{2}\subset \mathbb{R}$ are two subsets of points. We compute Chamfer distance by sampling uniformly $2048$ points from the predicted meshes and $100K$ points from the reference (this set is fixed for all the baselines).

\paragraph{Earth mover's distance (EMD).} The EMD is the solution of the optimization problem that aims at transforming one set to the other:
\begin{equation*}
d_{E M D}\left(S_{1}, S_{2}\right)=\min _{\phi: S_{1} \rightarrow S_{2}} \sum_{x \in S_{1}}\|x-\phi(x)\|_{2},
\end{equation*} 
where $\phi: S_{1} \rightarrow S_{2}$ is a bijection, and the two sets  $S_{1}, S_{2} \subseteq \mathbb{R}^{3}$ are of equal sizes. Due to the limited capacity of a GPU we sample $2048$ points from both the predicted meshes and the reference to compute this measure.

\paragraph{F-Score.} Tatarchenko et al.~\cite{tatarchenko2019single} observed that Chamfer distance is sensitive to outliers and proposed to use F-score for comparison of two point clouds. The F-score is defined as a harmonic mean between precision and recall. Precision and recall count the percentage of the points in one set for which there is a point in the other set within a distance threshold, set to 0.01 in our experiments. Note that the ground-truth shapes are normalized to have the largest dimension of 1, and the reconstruction results are globally aligned to the ground-truth.

\subsection{Style variance: Stylized dataset}
\label{sec:variance} 
\paragraph{Synthetic naive sketch vs stylized sketch} We first compare our baselines when trained on naive and stylized datasets on a chair category only. We evaluate the performance on the test set of shapes rendered with a naive style or stylized style, as well as on the corresponding sketches from ProSketch3D. 

\begin{figure}[ht]
\begin{center}
  \includegraphics[width=\linewidth]{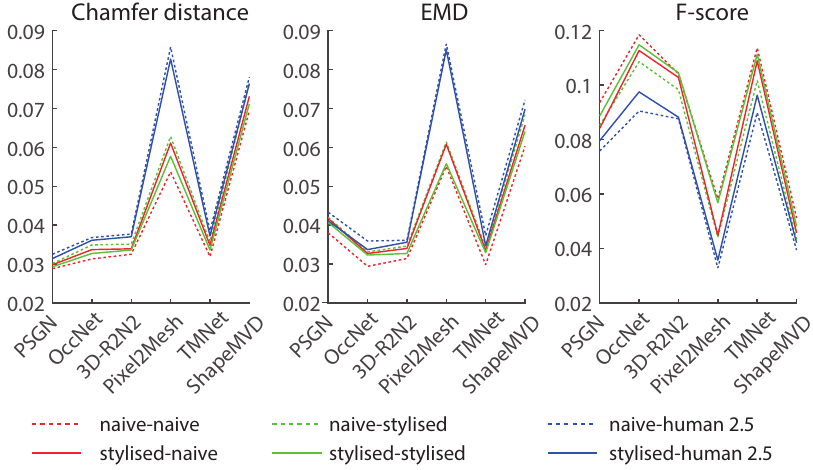}
\end{center}
  \caption{We train all the baselines with our naive and stylized datasets, and test on naive sketch test set, stylized dataset and sketches from SketchPro3D which matches our test set of shapes. In each pair, e.g. naive-naive, the first refers to the training set and the second to the test set. For the first two measures the lower is the better, while for the F-score the higher is the better.}
\label{fig:naive_stylised}
\vspace{-8pt}
\end{figure}

As can be seen in Figure \ref{fig:naive_stylised}, training on stylized sketches improves the performance on human sketches as judged by all measures. Yet, one can see that when trained with stylized sketches the performance on naive sketches is worse than when trained on naive sketches, implying that for the optimal performance the stylized dataset should include more clean sketch examples. Yet, training on stylized sketches and testing on naive sketches gives better results than training on naive and testing on stylized sketches, indicating the need of the stylization.

It can be observed that PSGN, OccNet, 3D-R2N2, TMNet and ShapeMVD are relatively robust to the different styles, while Pixel2Mesh shows bad generalization properties. Moreover, PSGN, OccNet, 3D-R2N2 and TMNet demonstrate comparable performance, leaving Pixel2Mesh and ShapeMVD behind.

\paragraph{Stylized sketch vs simplified sketch}
\begin{figure}[ht]
\begin{center}
   \includegraphics[width=\linewidth]{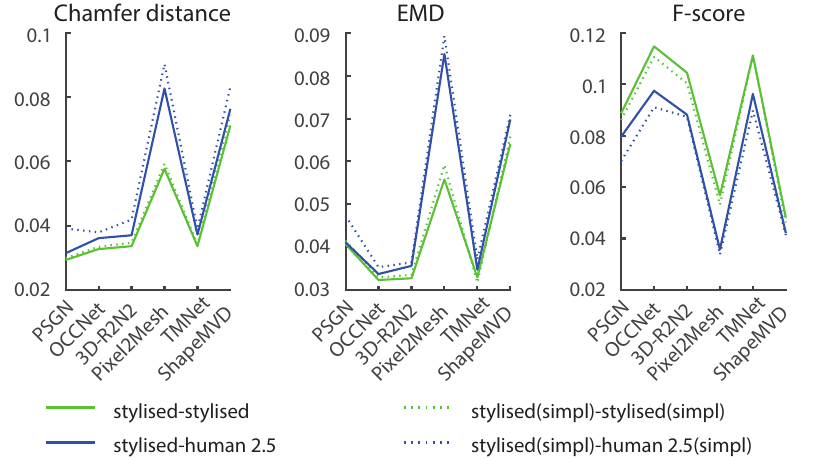}
\end{center}
   \caption{Comparison of the performance of the method when trained on our stylized dataset vs.~when trained with a usage of an additional sketch style-unifying network. We use the sketch simplification network by Simo-Serra et al.~\cite{simo2015discriminative}, denoted as `(simpl)'.}
\label{fig:stylised_vs_simplified}
\vspace{-4pt}
\end{figure}
We compare the performance of the methods trained on our stylized data and tested on stylized/human test sets of sketches, versus the performance of the methods trained on the stylized sketches processed with the simplifying network \cite{simo2015discriminative} and tested on the stylized/human sketches as well processed through this simplifying network. Figure \ref{fig:stylised_vs_simplified} shows that the performance of the methods trained with the stylized dataset outperforms the results obtained with a usage of a style-unifying (simplifying) network, indicating that, when one is interested in an accurate reconstruction using deep learning methods, the training data which models diverse sketching styles is preferable over a style-unifying network.

\paragraph{Single category vs Multiple categories.}
\begin{figure}[ht]
\vspace{-10pt}
\begin{center}
   \includegraphics[width=\linewidth]{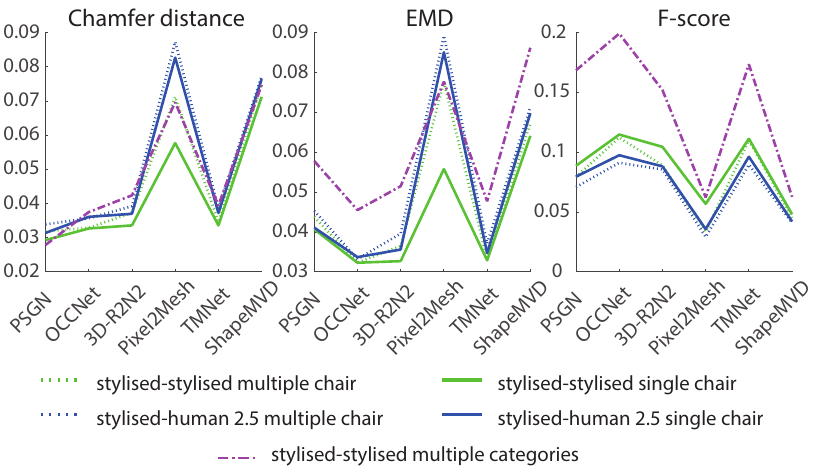}
\end{center}
   \caption{Comparison of models performance when trained on one category (chairs) at a time versus when trained on multiple categories: chairs, lamps and airplanes. 'multiple chairs' means that the training was done on multiple categories but tested only on chairs. 'multiple categories' means that training and testing are done on multiple categories. }
\label{fig:single_vs_multiple}
\vspace{-6pt}
\end{figure}
Training with multiple categories can be harder than training with a single category due to a large variance among categories. It can be seen in Figure \ref{fig:single_vs_multiple} that training on multiple categories and testing on a single chair category results in slightly worse performance on all baselines than training only on a single chair category, with an exception of TMNet where the results are almost identical. Interestingly, it can be seen that while Chamfer distance shows that the overall accuracy for multiple categories is comparable with a single chair category for all the baselines, the EMD measure shows that the accuracy of multiple categories is worse and F-score that it is better.

\subsection{Sparsity: Foreground mask}
\label{sec:sparsity}
\begin{figure}[htb]
\begin{center}
\includegraphics[width=\columnwidth]{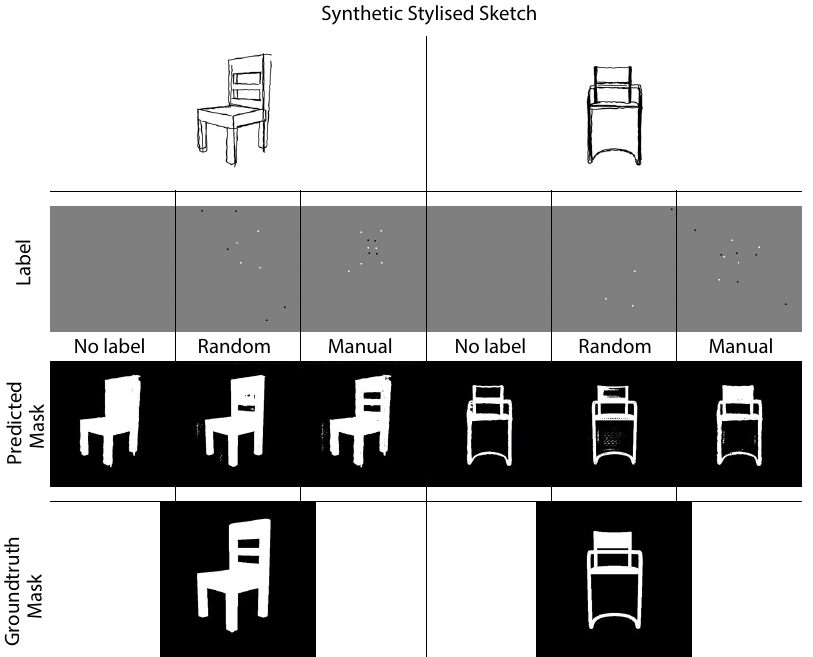}
\end{center}
   \caption{Example of mask predictions, without any label, with a randomly generated labeling and with human labeling.}
\label{fig:mask}
\vspace{-8pt}
\end{figure}

\paragraph{Evaluation of the foreground mask prediction}
Figure \ref{fig:mask} shows that with a few sparse user labels our foreground mask prediction network allows to achieve nearly perfect prediction of the foreground mask. Table \ref{tab:mask} provides numerical evaluation when no labels or automatically generated random labels are used. 

\begin{table}[ht]
\vspace{-2pt}
\centering
\small
\begin{tabular}{c|ccc}
\hline
Label         & IoU   & Precision & Recall \\ \hline
Random Label & 0.931 & 0.951     & 0.979  \\
No Label     & 0.834 & 0.887     & 0.892  \\ \hline
\end{tabular}
\vspace{2pt}
\caption{The evaluation of accuracy of the foreground mask prediction on a  synthetic stylized test dataset when trained on a stylized training set.}
\label{tab:mask}
\vspace{-12pt}
\end{table}

\begin{figure}[ht]
\begin{center}
   \includegraphics[width=\linewidth]{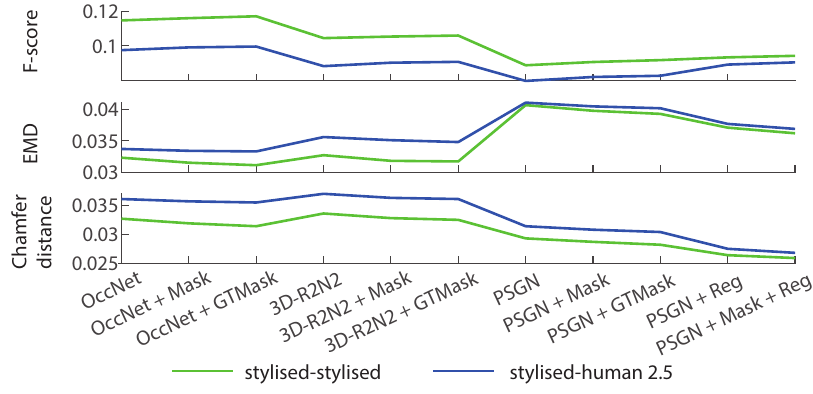}
\end{center}
   \caption{Evaluation of the effect of the proposed foreground mask branch and the regression loss. '+Mask' represents the result with the masks automatically obtained with our foreground prediction network with randomly generated labels, and '+GTMask' represents the results obtained with a ground-truth input mask, providing an upper estimate for the case when human labels are used. '+Reg' represents training with the proposed regression loss.}
\label{fig:final}
\vspace{-14pt}
\end{figure}

\begin{figure*}[htb]
\vspace{-8pt}
\begin{center}
\includegraphics[width=\linewidth]{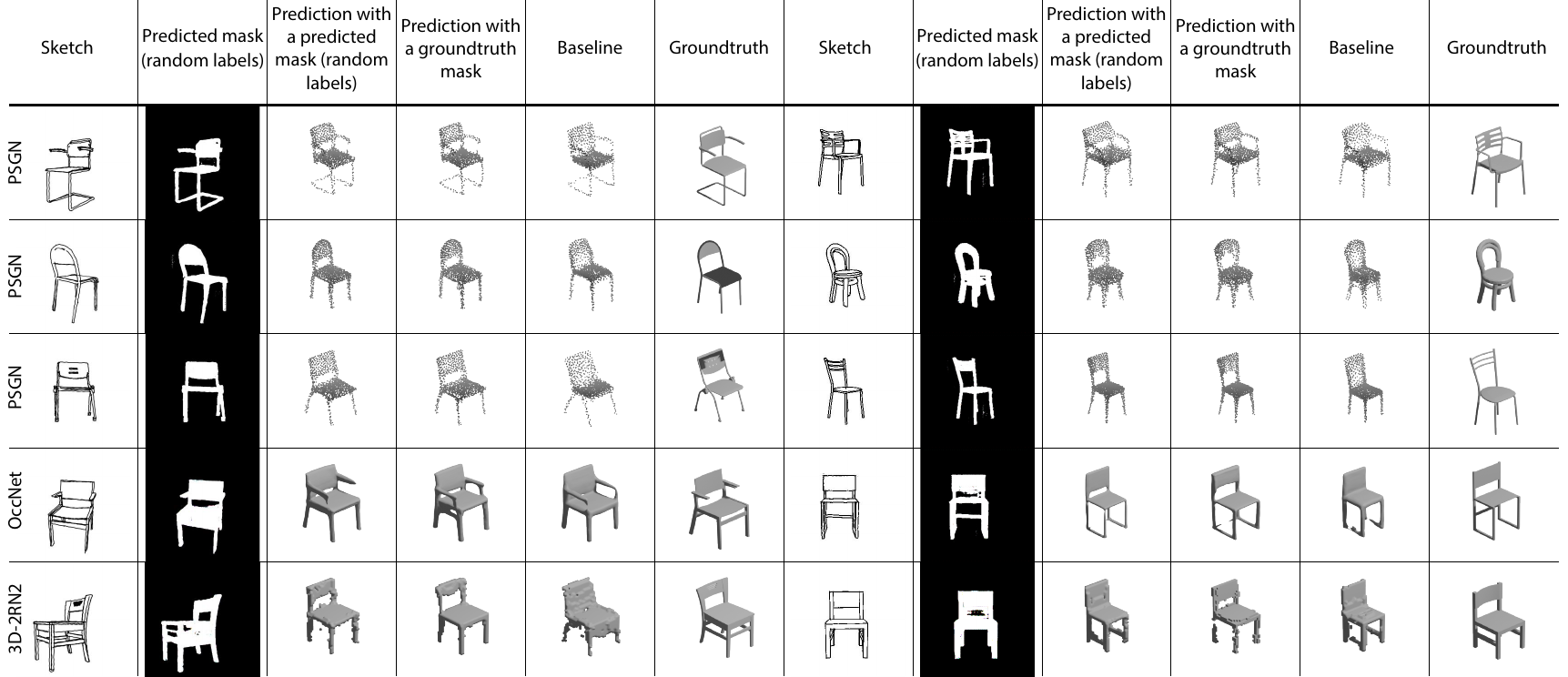}
\end{center}
\vspace{-2pt}
   \caption{Visual comparison of the reconstruction results with an additional input mask (predicted using randomly generated sparse labels or ground-truth) and without it.}
\label{fig:vis-mask}
\vspace{-8pt}
\end{figure*}

\paragraph{Training with a foreground mask}
For OccNet, PSGN and 3D-R2N2 baseline methods we train with a foreground mask as an additional input. Figure \ref{fig:final} shows that when the additional foreground mask is taken as an input the performance of all baselines is improved as judges by all three considered measures. 
Figure \ref{fig:vis-mask} shows qualitative improvements of the reconstruction results when an additional mask input is used.

\subsection{Style and view invariance: Regression loss}
\label{sec:view}
\begin{figure}[htb!]
\vspace{-12pt}
\begin{center}
\includegraphics[width=\linewidth]{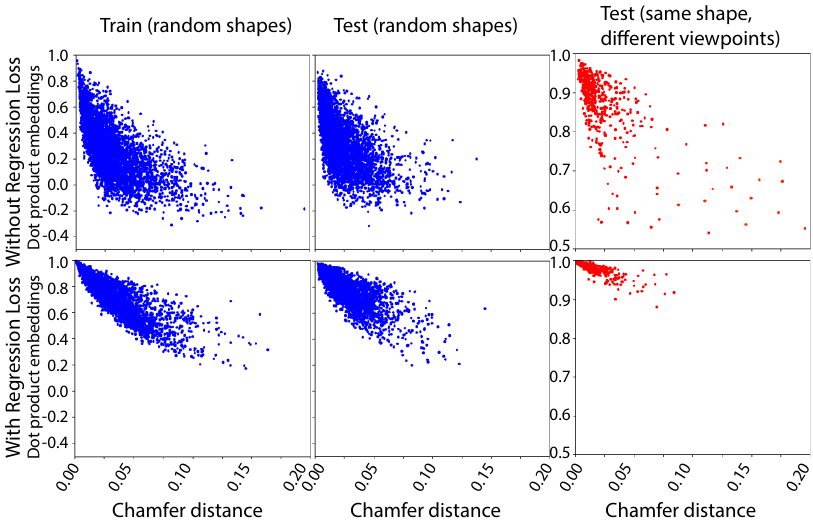}
\end{center}
   \caption{Comparison of latent space distributions. The y-axes is a dot product in the space of sketches embedding, and x-axes is a Chamfer distance between the corresponding 3D shapes. The blue corresponds to pairs of random shapes. For the red for each shape in a test set we fix a viewpoint with $45^\circ$ azimuth as a reference. We then compute the distances from the reference view embedding to the embeddings of all other viewpoints of the same shape, and Chamfer distances from the reference viewpoint reconstruction to the reconstructions from other viewpoints.}
\label{fig:latent_space}
\vspace{-12pt}
\end{figure}

We train PSGN with the combination of the reconstruction and the regression losses. Note that any of the volumetric baselines apart from OccNet, which does not have an explicit 3D shape representation at the training stage, can be trained with the proposed regression loss. Figure \ref{fig:final} shows that the proposed regression loss boosts the performance, where combining it with the foreground mask input allows network to achieve the optimal performance among all PSGN-based baselines. We plot in Figure \ref{fig:latent_space} how Chamfer distances correlate with dot product distance in the feature space. It shows that the regression loss pushes the embeddings of the sketches of similar shapes further apart, allowing to learn a more descriptive feature space (blue). It also shows that the regression loss makes the reconstructions among viewpoints with different styles significantly more consistent (red).

\section{Conclusion}
In this work instead of focusing on a particular shape representation or a network architecture, we focus on key differences between sparse sketches and 2D images. We demonstrate how an additional network can be used to predict foreground mask from an input sketch and sparse human labels, we show that when such a mask is passed alongside a sketch to a reconstruction network the performance increases across all considered baselines. We introduced a regression loss which allows to learn a more discriminate embedding of input sketches. We believe that our work will serve as a reference for deep sketch-based modeling and will encourage future development of dedicated reconstruction networks that take sketch specifics into account. All the datasets and models are available at \url{https://tinyurl.com/DeepSketchModeling}.

{\small
\bibliographystyle{ieee}
\bibliography{literature_sketch}
}

\end{document}